\documentclass{article} 
\usepackage[final]{nips_2016}
\usepackage{hyperref}
\usepackage{url}
\usepackage{booktabs}
\usepackage{graphicx}
\usepackage{amsmath,amssymb}
\usepackage{array}
\usepackage{multirow}
\usepackage[usenames, dvipsnames]{color}
\usepackage{colortbl,soul}
\usepackage[compact]{titlesec}
\usepackage{enumitem}
\usepackage{appendix}
\usepackage{multicol}
\usepackage{graphicx} 
\usepackage{subfigure} 
\usepackage{natbib}
\usepackage{hyperref}
\usepackage{amsmath}
\usepackage{amsfonts}
\usepackage{bibunits}
\usepackage{algorithm}
\usepackage{algpseudocode}
\usepackage{algorithmicx}
\usepackage{xspace}

\newcommand{\figref}[1]{Fig.~\ref{#1}}

\newcommand{\sref}[1]{Sect.~\ref{#1}}
\def\eg{\emph{e.g.}}
\def\ie{\emph{i.e.}}

\def\mainobj{$\mathcal{Q}$}
\def\diffobj{$\hat{\mathcal{Q}}$}

\newcommand{\pbtexploit}{\texttt{exploit}\xspace}
\newcommand{\pbtexplore}{\texttt{explore}\xspace}
\newcommand{\pbtstep}{\texttt{step}\xspace}
\newcommand{\pbteval}{\texttt{eval}\xspace}

\definecolor{Gray}{gray}{0.95}

\DeclareMathOperator*{\argmax}{arg\,max}

\newcommand{\thetam}{\theta}
\newcommand{\thetao}{h}

\title{Population Based Training of Neural Networks}

\author{
Max Jaderberg \\
\And
Valentin Dalibard \\
\And
Simon Osindero \\
\And
Wojciech M. Czarnecki \\
\AND
Jeff Donahue \\
\And
Ali Razavi \\
\And
Oriol Vinyals \\
\And
Tim Green \\
\And
Iain Dunning\\
\AND
Karen Simonyan \\
\And
Chrisantha Fernando \\
\And
Koray Kavukcuoglu
}

\begin{document}

\maketitle

\vspace{-3.0em}
 \begin{center}
 DeepMind, London, UK
 \end{center}
 \vspace{1em}

\begin{abstract}

Neural networks dominate the modern machine learning landscape, but their training and success still suffer from sensitivity to empirical choices of hyperparameters such as model architecture, loss function, and optimisation algorithm.
In this work we present \emph{Population Based Training (PBT)}, a simple asynchronous optimisation algorithm which effectively utilises a fixed computational budget to jointly optimise a population of models and their hyperparameters to maximise performance.
Importantly, PBT discovers a schedule of hyperparameter settings rather than following the generally sub-optimal strategy of trying to find a single fixed set to use for the whole course of training.
With just a small modification to a typical distributed hyperparameter training framework, our method allows robust and reliable training of models. 
We demonstrate the effectiveness of PBT on deep reinforcement learning problems, showing faster wall-clock convergence and higher final performance of agents by optimising over a suite of hyperparameters. 
In addition, we show the same method can be applied to supervised learning for machine translation, where PBT is used to maximise the BLEU score directly, and also to training of Generative Adversarial Networks to maximise the Inception score of generated images.
In all cases PBT results in the automatic discovery of hyperparameter schedules and model selection which results in stable training and better final performance.

\end{abstract}

\section{Introduction}

Neural networks have become the workhorse non-linear function approximator in a number of machine learning domains, most notably leading to significant advances in reinforcement learning (RL) and supervised learning. 
However, it is often overlooked that the success of a particular neural network model depends upon the joint tuning of the model structure, the data presented, and the details of how the model is optimised. Each of these components of a learning framework is controlled by a number of parameters, \emph{hyperparameters}, which influence the learning process and must be properly tuned to fully unlock the network performance.
This essential tuning process is computationally expensive, and as neural network systems become more complicated and endowed with even more hyperparameters, the burden of this search process becomes increasingly heavy. 
Furthermore, there is yet another level of complexity in scenarios such as deep reinforcement learning, where the learning problem itself can be highly non-stationary (\eg~dependent on which parts of an environment an agent is currently able to explore). As a consequence, it might be the case that the ideal hyperparameters for such learning problems are themselves highly non-stationary, and should vary in a way that precludes setting their schedule in advance.

Two common tracks for the tuning of hyperparameters exist: \emph{parallel search} and \emph{sequential optimisation}, which trade-off concurrently used computational resources with the time required to achieve optimal results. 
Parallel search performs many parallel optimisation processes (by \emph{optimisation process} we refer to neural network training runs), each with different hyperparameters, with a view to finding a single best output from one of the optimisation processes -- examples of this are grid search and random search. 
Sequential optimisation performs few optimisation processes in parallel, but does so many times sequentially, to gradually perform hyperparameter optimisation using information obtained from earlier training runs to inform later ones -- examples of this are hand tuning and Bayesian optimisation. Sequential optimisation will in general provide the best solutions, but requires multiple sequential training runs, which is often unfeasible for lengthy optimisation processes.

In this work, we present a simple method, Population Based Training (PBT) which bridges and extends parallel search methods and sequential optimisation methods.
Advantageously, our proposal has a wall-clock run time that is no greater than that of a single optimisation process, does not require sequential runs, and is also able to use fewer computational resources than naive search methods such as random or grid search.
Our approach leverages information sharing across a population of concurrently running optimisation processes, and allows for online propagation/transfer of parameters and hyperparameters between members of the population based on their performance. Furthermore, unlike most other adaptation schemes, our method is capable of performing online adaptation of hyperparameters -- which can be particularly important in problems with highly non-stationary learning dynamics, such as reinforcement learning settings.
PBT is decentralised and asynchronous, requiring minimal overhead and infrastructure. While inherently greedy, we show that this meta-optimisation process results in effective and automatic tuning of hyperparameters, allowing them to be adaptive throughout training. In addition, the model selection and propagation process ensures that intermediate good models are given more computational resources, and are used as a basis of further optimisation and hyperparameter search. 

We apply PBT to a diverse set of problems and domains to demonstrate its effectiveness and wide applicability. Firstly, we look at the problem of deep reinforcement learning, showing how PBT can be used to optimise UNREAL~\citep{jaderberg2016reinforcement} on DeepMind Lab levels~\citep{beattie2016deepmind}, Feudal Networks~\citep{vezhnevets2017feudal} on Atari games~\citep{bellemare2013arcade}, and simple A3C agents for StarCraft II~\citep{vinyals2017starcraft}. In all three cases we show faster learning and higher performance across a suite of tasks, with PBT allowing discovery of new state-of-the-art performance and behaviour, as well as the potential to reduce the computational resources for training. Secondly, we look at its use in supervised learning for machine translation on WMT 2014 English-to-German with Transformer networks~\citep{vaswani2017attention}, showing that PBT can match and even outperform heavily tuned hyperparameter schedules by optimising for BLEU score directly. Finally, we apply PBT to the training of Generative Adversarial Networks (GANs)~\citep{goodfellowgan} by optimising for the Inception score~\citep{openaigan} -- the result is stable GAN training with
large performance improvements over a strong, well-tuned baseline with the same architecture~\citep{dcgan}.

The improvements we show empirically are the result of (a) automatic selection of hyperparameters during training,  (b) online model selection to maximise the use of computation spent on promising models, and (c) the ability for online adaptation of hyperparameters to enable non-stationary training regimes and the discovery of complex hyperparameter schedules. 

In \sref{sec:related} we review related work on parallel and sequential hyperparameter optimisation techniques, as well as evolutionary optimisation methods which bear a resemblance to PBT. We introduce PBT in \sref{sec:pbt}, outlining the algorithm in a general manner which is used as a basis for experiments. The specific incarnations of PBT and the results of experiments in different domains are given in \sref{sec:exp}, and finally we conclude in \sref{sec:conclusion}.

\begin{figure}[t]
\centering
\includegraphics[width=1.0\textwidth]{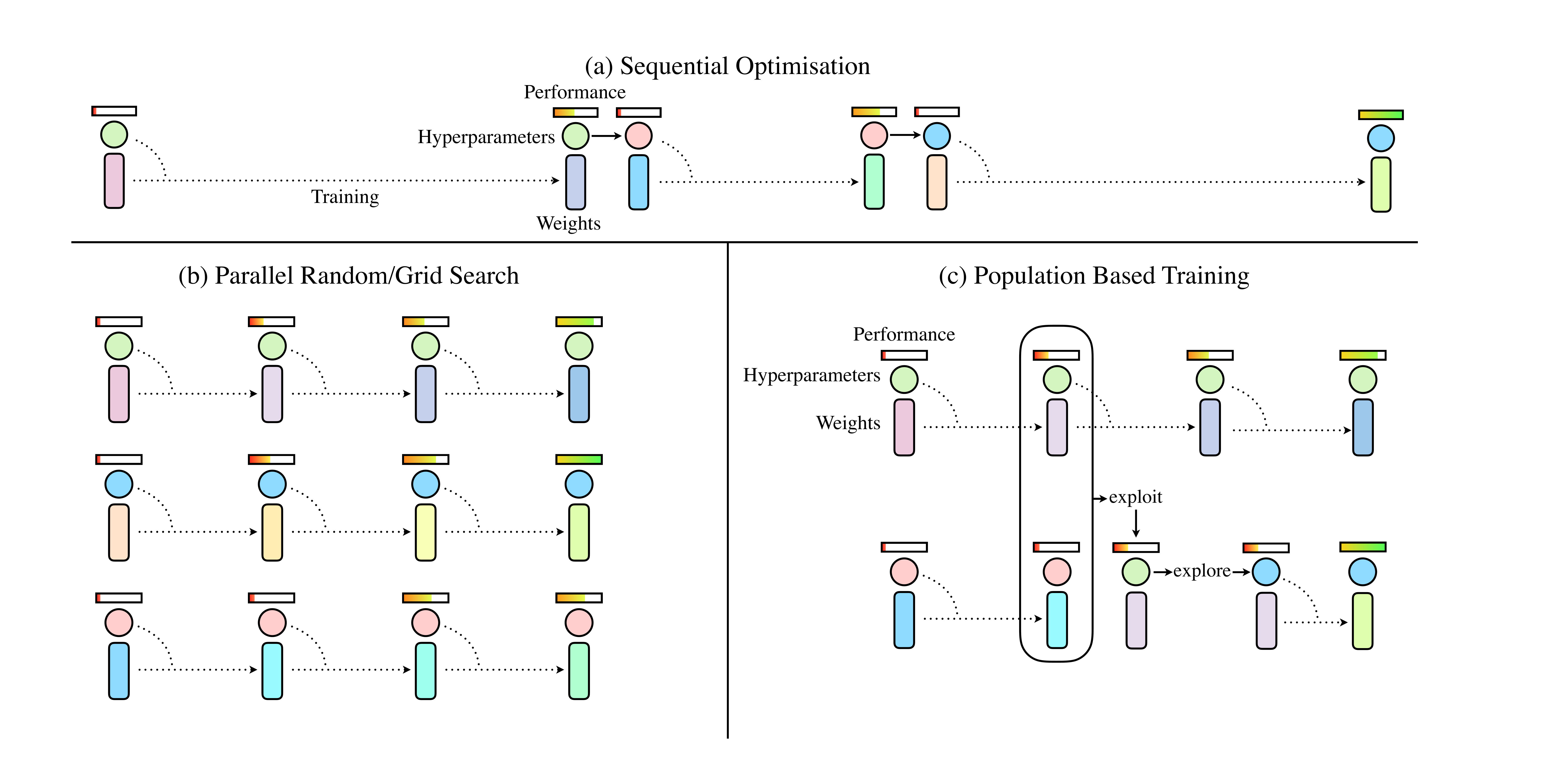}
\caption{\small Common paradigms of tuning hyperparameters: \emph{sequential optimisation} and \emph{parallel search}, as compared to the method of \emph{population based training} introduced in this work. (a) Sequential optimisation requires multiple training runs to be completed (potentially with early stopping), after which new hyperparameters are selected and the model is retrained from scratch with the new hyperparameters. This is an inherently sequential process and leads to long hyperparameter optimisation times, though uses minimal computational resources. (b) Parallel random/grid search of hyperparameters trains multiple models in parallel with different weight initialisations and hyperparameters, with the view that one of the models will be optimised the best. This only requires the time for one training run, but requires the use of more computational resources to train many models in parallel. (c) Population based training starts like parallel search, randomly sampling hyperparameters and weight initialisations. However, each training run asynchronously evaluates its performance periodically. If a model in the population is under-performing, it will \emph{exploit} the rest of the population by replacing itself with a better performing model, and it will \emph{explore} new hyperparameters by modifying the better model's hyperparameters, before training is continued. This process allows hyperparameters to be optimised online, and the computational resources to be focused on the hyperparameter and weight space that has most chance of producing good results. The result is a hyperparameter tuning method that while very simple, results in faster learning, lower computational resources, and often better solutions.}
\label{fig:overview}
\end{figure}

\section{Related Work}\label{sec:related}

First, we review the methods for sequential optimisation and parallel search for hyperparameter optimisation that go beyond grid search, random search, and hand tuning. Next, we note that PBT inherits many ideas from genetic algorithms, and so highlight the literature in this space.

The majority of automatic hyperparameter tuning mechanisms are sequential optimisation methods: the result of each training run with a particular set of hyperparameters is used as knowledge to inform the subsequent hyperparameters searched. A lot of previous work uses a Bayesian optimisation framework to incorporate this information by updating the posterior of a Bayesian model of successful hyperparameters for training -- examples include GP-UCB~\citep{srinivas2009gaussian}, TPE~\citep{bergstra2011algorithms}, Spearmint~\citep{snoek2012practical}, and SMAC~\citep{hutter2011sequential}. As noted in the previous section, their sequential nature makes these methods prohibitively slow for expensive optimisation processes. For this reason, a number of methods look to speed up the updating of the posterior with information from each optimisation process by performing early stopping, using intermediate losses to predict final performance, and even modelling the entire time and data dependent optimisation process in order to reduce the number of optimisation steps required for each optimisation process and to better explore the space of promising hyperparameters~\citep{gyorgy2011efficient,agarwal2011oracle,sabharwal2016selecting,swersky2013multi,swersky2014freeze,domhan2015speeding,klein2016fast,snoek2015scalable,springenberg2016bayesian}. 

Many of these Bayesian optimisation approaches can be further sped up by attempting to parallelise them -- training independent models in parallel to quicker update the Bayesian model (though potentially introducing bias)~\citep{shah2015parallel,gonzalez2016batch,wu2016parallel,rasley2017hyperdrive,golovin2017google} -- but these will still require multiple sequential model optimisations. The same problem is also encountered where genetic algorithms are used in place of Bayesian optimisation for hyperparameter evolution~\citep{young2015optimizing}.

Moving away from Bayesian optimisation, even approaches such as Hyperband~\citep{li2016hyperband} which model the problem of hyperparameter selection as a many-armed bandit, and natively incorporate parallelisation, cannot practically be executed within a single training optimisation process due to the large amount of computational resources they would initially require.

The method we present in this paper only requires a single training optimisation process -- it builds upon the unreasonable success of random hyperparameter search which has shown to be very effective~\citep{bergstra2012random}. Random search is effective at finding good regions for sensitive hyperparameters -- PBT identifies these and ensures that these areas are explored more using partially trained models (\eg~by copying their weights). By bootstrapping the evaluation of new hyperparameters during training on partially trained models we eliminate the need for sequential optimisation processes that plagues the methods previously discussed. PBT also allows for adaptive hyperparameters during training, which has been shown, especially for learning rates, to improve optimisation~\citep{loshchilov2016sgdr,smith2017cyclical,masse2015speed}. 

PBT is perhaps most analogous to evolutionary strategies that employ self-adaptive hyperparameter tuning to modify how the genetic algorithm itself operates~\citep{back1998overview,Spears95adaptingcrossover,gloger2004self,clune2008natural} -- in these works, the evolutionary hyperparameters were mutated at a slower rate than the parameters themselves, similar to how hyperparameters in PBT are adjusted at a slower rate than the parameters. In our case however, rather than evolving the parameters of the model, we train them partially with standard optimisation techniques (\eg~gradient descent). Other similar works to ours are Lamarckian evolutionary algorithms in which parameters are inherited whilst hyperparameters are evolved \citep{castillo2006lamarckian} or where the hyperparameters, initial weights, and architecture of networks are evolved, but the parameters are learned \citep{castillo1999g,husken2000optimization}, or where evolution and learning are both applied to the parameters \citep{ku1997exploring,houck1997empirical,igel2005evolutionary}. This mixture of learning and evolutionary-like algorithms has also been explored successfully in other domains~\citep{zhang2011evolutionary} such as neural architecture search~\citep{real2017large,liu2017hierarchical}, feature selection~\citep{xue2016survey}, and parameter learning~\citep{fernando2016convolution}. Finally, there are parallels to work such as \cite{salustowicz1997probabilistic} which perform program search with genetic algorithms.

\section{Population Based Training}\label{sec:pbt}

The most common formulation in machine learning is to optimise the parameters $\thetam$ of a model $f$ to maximise a given objective function \diffobj, (\eg~classification, reconstruction, or prediction). Generally, the trainable parameters $\thetam$ are updated using an optimisation procedure such as stochastic gradient descent.
More importantly, the actual performance metric \mainobj{} that we truly care to optimise is often different to \diffobj{}, for example \mainobj{} could be accuracy on a validation set, or BLEU score as used in machine translation. The main purpose of PBT is to provide a way to optimise both the parameters $\thetam$ and the hyperparameters $\thetao$ jointly on the actual metric \mainobj{} that we care about.

To that end, firstly we define a function \pbteval that evaluates the objective function, \mainobj{}, using the current state of model $f$. For simplicity we ignore all terms except $\thetam$, and define \pbteval as function of trainable parameters $\thetam$ only. The process of finding the optimal set of parameters that maximise \mainobj{} is then:
\begin{equation}
\theta^* = \argmax_{\theta \in \Theta} \mathtt{eval}(\theta).
\label{eqn:optimise}
\end{equation}

Note that in the methods we are proposing, the \pbteval function does not need to be differentiable, nor does it need to be the same as the function used to compute the iterative updates in the optimisation steps which we outline below (although they ought to be correlated).

When our model is a neural network, we generally optimise the weights $\thetam$ in an iterative manner, \eg~by using stochastic gradient descent on the objective function $\mathcal{Q}$. Each step of this iterative optimisation procedure, \pbtstep, updates the parameters of the model, and is itself conditioned on some parameters $\thetao \in \mathcal{H}$ (often referred to as hyperparameters). 

In more detail, iterations of parameter update steps:
\begin{equation}
\thetam \leftarrow \mathtt{step}(\thetam | \thetao)
\label{eqn:step}
\end{equation}
are chained to form a sequence of updates that ideally converges to the optimal solution
\begin{equation}
\thetam^* =  \mathtt{optimise}(\thetam | \boldsymbol{\thetao} ) = \mathtt{optimise}(\thetam | (\thetao_t )_{t=1}^T) = \mathtt{step}(\mathtt{step}(\ldots\mathtt{step}(\thetam | \thetao_1)\ldots | \thetao_{T-1})| \thetao_T).
\end{equation}
This iterative optimisation process can be computationally expensive, due to the number of steps $T$ required to find $\thetam^*$ as well as the computational cost of each individual step, often resulting in the optimisation of $\thetam$ taking days, weeks, or even months. In addition, the solution is typically very sensitive to the choice of the sequence of hyperparemeters $\boldsymbol{\thetao} = (\thetao_t )_{t=1}^T$. 
Incorrectly chosen hyperparameters can lead to bad solutions or even a failure of the optimisation of $\thetam$ to converge.
Correctly selecting hyperparameters requires strong prior knowledge on $\boldsymbol{\thetao}$ to exist or to be found (most often through multiple optimisation processes with different $\boldsymbol{\thetao}$).
Furthermore, due to the dependence of $\boldsymbol{\thetao}$ on iteration step, the number of possible values grows exponentially with time. Consequently, it is common practise to either make all $\thetao_t$ equal to each other (\eg~constant learning rate through entire training, or constant regularisation strength) or to predefine a simple schedule (\eg~learning rate annealing). 
In both cases one needs to search over multiple possible values of $\boldsymbol{\thetao}$
\begin{equation}
\thetam^* = \mathtt{optimise}(\thetam | \boldsymbol{\thetao}^*), ~\boldsymbol{\thetao}^* = \argmax_{\boldsymbol{\thetao} \in \mathcal{H}^T} \mathtt{eval}(\mathtt{optimise}(\thetam | \boldsymbol{\thetao})).
\label{eqn:machinelearning}
\end{equation}

As a way to perform \eqref{eqn:machinelearning} in a fast and computationally efficient manner, we consider training $N$ models $\{\thetam^{i}\}_{i=1}^N$ forming a \emph{population} $\mathcal{P}$ which are optimised with different hyperparameters $\{\boldsymbol{\thetao}^{i}\}_{i=1}^N$. The objective is to therefore find the optimal model across the entire population. However, rather than taking the approach of parallel search, we propose to use the collection of partial solutions in the population to additionally perform meta-optimisation, where the hyperparameters $\thetao$ and weights $\thetam$ are additionally adapted according to the performance of the entire population. 

In order to achieve this, PBT uses two methods called independently on each member of the population (each worker): \pbtexploit, which, given performance of the whole population, can decide whether the worker should abandon the current solution and instead focus on a more promising one; and \pbtexplore, which given the current solution and hyperparameters proposes new ones to better explore the solution space.

\begin{algorithm}[t]
    \caption{Population Based Training (PBT)}
    \label{pbtalg}
    \begin{algorithmic}[1] 
        \Procedure{Train}{$\mathcal{P}$} \Comment{initial population $\mathcal{P}$}
            \For{($\theta$, $h$, $p$, $t$) $\in \mathcal{P}$ (asynchronously in parallel)}
            \While{not end of training} 
                \State $\theta \gets \mathtt{step}(\theta|h)$ \Comment{one step of optimisation using hyperparameters $h$}
                \State $p \gets \mathtt{eval}(\theta)$ \Comment{current model evaluation}
                \If{$\mathtt{ready}(p,t,\mathcal{P})$}
                \State $h', \theta' \gets \mathtt{exploit}(h,\theta,p,\mathcal{P})$\Comment{use the rest of population to find better solution}
                \If{$\theta \neq \theta'$}
                \State $h, \theta \gets \mathtt{explore}(h', \theta', \mathcal{P})$ \Comment{produce new hyperparameters $h$}
                \State $p \gets \mathtt{eval}(\theta)$  \Comment{new model evaluation}
                \EndIf
                \EndIf
            \State update $\mathcal{P}$ with new $(\theta, h, p, t + 1)$ \Comment{update population}
            \EndWhile\label{euclidendwhile}
            \EndFor
            \State \textbf{return} $\theta$ with the highest $p$ in $\mathcal{P}$
        \EndProcedure
    \end{algorithmic}
\label{alg:pbt}
\end{algorithm}

\begin{figure}
\centering
\includegraphics[width=0.4\textwidth]{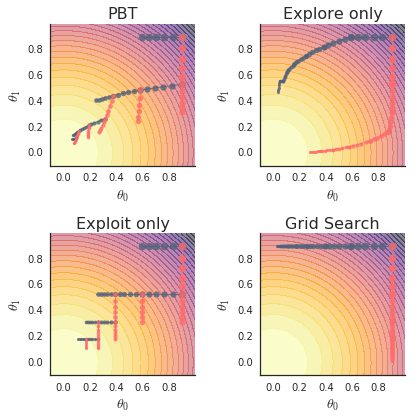}
\includegraphics[width=0.4\textwidth]{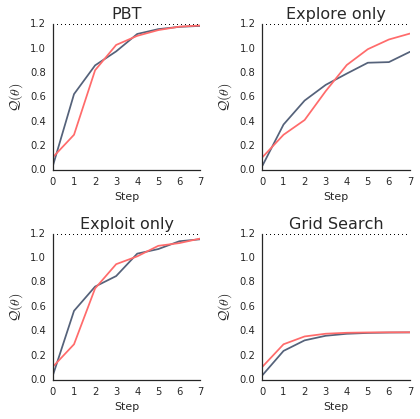}
\caption{\small A visualisation of PBT on a toy example where our objective is to maximise a simple quadratic function \mainobj$(\theta) = 1.2 - (\theta_0^2 + \theta_1^2)$, but without knowing its formula. Instead, we are given a surrogate function \diffobj$(\theta | h) = 1.2 - ( h_0 \theta_0^2 + h_1 \theta_1^2)$ that we can differentiate. This is a very simple example of the setting where one wants to maximise one metric (\eg~generalisation capabilities of a model) while only being able to directly optimise other one (\eg~empirical loss). We can, however, read out the values of \mainobj{} for evaluation purposes. We can treat $h$ as hyperparameters, which can change during optimisation, and use gradient descent to minimise $Q(\theta | h)$. For simplicity, we assume there are only two workers (so we can only run two full maximisations), and that we are given initial $\theta_0 = [0.9, 0.9]$. With grid/random search style approaches we can only test two values of $h$, thus we choose $[1, 0]$ (black), and $[0, 1]$ (red). Note, that if we would choose $[1, 1]$ we would recover true objective, but we are assuming that we do not know that in this experiment, and in general there is no $h$ such that \diffobj$(\theta|h)= $ \mainobj$(\theta)$.
As one can see, we increase \mainobj{} in each step of maximising $Q(\cdot|h)$, but end up far from the actual optimum (\mainobj$(\theta) \approx 0.4$), due to very limited exploration of hyperparameters $h$. 
If we apply PBT to the same problem where every 4 iterations the weaker of the two workers copies the solution of the better one (exploitation) and then slightly perturbs its update direction (exploration), we converge to a global optimum. 
Two ablation results show what happens if we only use exploration or only exploitation. We see that majority of the performance boost comes from copying solutions between workers (exploitation), and exploration only provides additional small improvement. Similar results are obtained in the experiments on real problems, presented in \sref{sec:analysis}.
\textit{Left}: The learning over time, each dot represents a single solution and its size decreases with iteration. \textit{Right}: The objective function value of each worker over time.}
\label{fig:basic}
\end{figure}

Each member of the population is trained in parallel, with iterative calls to \pbtstep to update the member's weights and \pbteval to measure the member's current performance. However, when a member of the population is deemed ready (for example, by having been optimised for a minimum number of steps or having reached a certain performance threshold), its weights and hyperparameters are updated by \pbtexploit and \pbtexplore. For example, \pbtexploit could replace the current weights with the weights that have the highest recorded performance in the rest of the population, and \pbtexplore could randomly perturb the hyperparameters with noise. After \pbtexploit and \pbtexplore, iterative training continues using \pbtstep as before. This cycle of local iterative training (with \pbtstep) and exploitation and exploration using the rest of the population (with \pbtexploit and \pbtexplore) is repeated until convergence of the model. Algorithm~\ref{alg:pbt} describes this approach in more detail, \figref{fig:overview} schematically illustrates this process (and contrasts it with sequential optimisation and parallel search), and \figref{fig:basic} shows a toy example illustrating the efficacy of PBT.

The specific form of \pbtexploit and \pbtexplore depends on the application. In this work we focus on optimising neural networks for reinforcement learning, supervised learning, and generative modelling with PBT (\sref{sec:exp}).
In these cases, \pbtstep is a step of gradient descent (with \eg~SGD or RMSProp~\citep{tieleman2012lecture}), \pbteval is the mean episodic return or validation set performance of the metric we aim to optimise, \pbtexploit selects another member of the population to copy the weights and hyperparameters from, and \pbtexplore creates new hyperparameters for the next steps of gradient-based learning by either perturbing the copied hyperparameters or resampling hyperparameters from the originally defined prior distribution. A member of the population is deemed ready to exploit and explore when it has been trained with gradient descent for a number of steps since the last change to the hyperparameters, such that the number of steps is large enough to allow significant gradient-based learning to have occurred.

By combining multiple steps of gradient descent followed by weight copying by \pbtexploit, and perturbation of hyperparameters by \pbtexplore, we obtain learning algorithms which benefit from not only local optimisation by gradient descent, but also periodic model selection, and hyperparameter refinement from a process that is more similar to genetic algorithms, creating a two-timescale learning system.
An important property of population based training is that it is asynchronous and does not require a centralised process to orchestrate the training of the members of the population. Only the current performance information, weights, and hyperparameters must be globally available for each population member to access -- crucially there is no synchronisation of the population required.

\section{Experiments}\label{sec:exp}

\begin{figure}[t]
\centering
\begin{tabular}{cccc}
\includegraphics[width=1.0\textwidth]{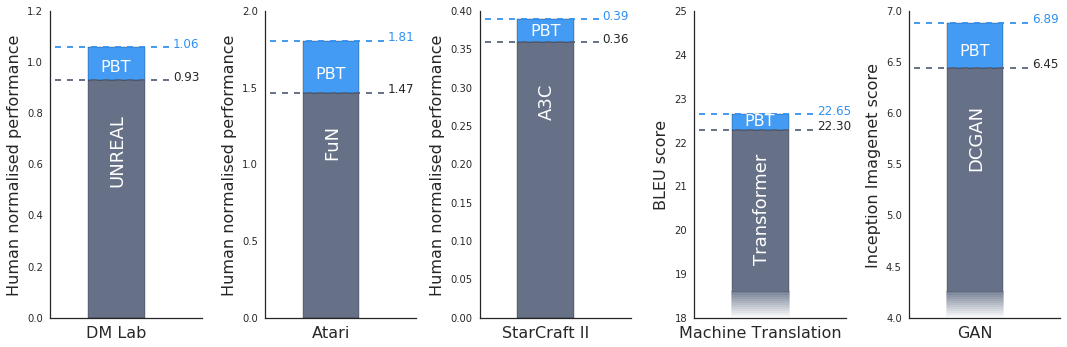}
\end{tabular}
\caption{\small The results of using PBT over random search on different domains. \emph{DM Lab:} We show results for training UNREAL with and without PBT, using a population of 40 workers, on 15 DeepMind Lab levels, and report the final human-normalised performance of the best agent in the population averaged over all levels. \emph{Atari:} We show results for Feudal Networks with and without PBT, using a population of 80 workers on each of four games: Amidar, Gravitar, Ms-Pacman, and Enduro, and report the final human-normalised performance of the best agent in the population averaged over all games. \emph{StarCraft II:} We show the results for training A3C with and without PBT, using a population of 30 workers, on a suite of 6 mini-game levels, and report the final human-normalised performance of the best agent in the population averaged over all levels. \emph{Machine Translation:} We show results of training Transformer networks for machine translation on WMT 2014 English-to-German. We use a population of 32 workers with and without PBT, optimising the population to maximise BLEU score. \emph{GAN:} We show results for training GANs with and without PBT using a population size of 45, optimising for Inception score. Full results on all levels can be found in \sref{sec:detailed_results_RL}.
}
\label{fig:showstopper}
\end{figure}

In this section we will apply Population Based Training to different learning problems. We describe the specific form of PBT for deep reinforcement learning in \sref{sec:deeprl} when applied to optimising UNREAL~\citep{jaderberg2016reinforcement} on DeepMind Lab 3D environment tasks~\citep{beattie2016deepmind}, Feudal Networks~\citep{vezhnevets2017feudal} on the Atari Learning Environment games~\citep{bellemare2013arcade}, and the StarCraft II environment baseline agents~\citep{vinyals2017starcraft}. In \sref{sec:language} we apply PBT to optimising state-of-the-art language models, transformer networks~\citep{vaswani2017attention}, for machine translation task. Next, in \sref{sec:gans} we apply PBT to the optimisation of Generative Adversarial Networks~\citep{goodfellowgan}, a notoriously unstable optimisation problem. In all these domains we aim to build upon the strongest baselines, and show improvements in training these state-of-the-art models by using PBT. Finally, we analyse the design space of Population Based Training with respect to this rich set of experimental results to draw practical guidelines for further use in \sref{sec:analysis}.

\subsection{Deep Reinforcement Learning}\label{sec:deeprl}
In this section we apply Population Based Training to the training of neural network agents with reinforcement learning (RL)
where we aim to find a policy $\pi$ to maximise expected episodic return $\mathbb{E}_\pi[R]$ within an environment. 
We first focus on A3C-style methods~\cite{mnih2016asynchronous} to perform this maximisation, along with some state-of-the-art extensions: UNREAL~\citep{jaderberg2016reinforcement} and Feudal Networks (FuN)~\citep{vezhnevets2017feudal}.

These algorithms are sensitive to hyperparameters, which include learning rate, entropy cost, and auxiliary loss weights, as well as being sensitive to the reinforcement learning optimisation process which can suffer from local minima or collapse due to the insufficient or unlucky policy exploration. PBT therefore aims to stabilise this process.

\subsubsection{PBT for RL}\label{sec:pbtforrl}

We use population sizes between 10 and 80, where each member of the population is itself a distributed asynchronous actor critic (A3C) style agent~\citep{mnih2016asynchronous}. 
\vspace{-0.5em}\paragraph{Hyperparameters}We allow PBT to optimise the learning rate, entropy cost, and unroll length for UNREAL on DeepMind Lab, learning rate, entropy cost, and intrinsic reward cost for FuN on Atari, and learning rate only for A3C on StarCraft~II. 
\vspace{-0.5em}\paragraph{Step}Each iteration does a step of gradient descent with RMSProp~\citep{tieleman2012lecture} on the model weights, with the gradient update provided by vanilla A3C, UNREAL or FuN.
\vspace{-0.5em}\paragraph{Eval}We evaluate the current model with the last 10 episodic rewards during training.
\vspace{-0.5em}\paragraph{Ready} A member of the population is deemed ready to go through the exploit-and-explore process using the rest of the population when between $1\times10^6$ to $10\times10^6$ agent steps have elapsed since the last time that population member was ready.
\vspace{-0.5em}\paragraph{Exploit}We consider two exploitation strategies. (a) \emph{T-test selection} where we uniformly sample another agent in the population, and compare the last 10 episodic rewards using Welch's t-test~\citep{welch1947generalization}. If the sampled agent has a higher mean episodic reward and satisfies the t-test, the weights and hyperparameters are copied to replace the current agent. (b) \emph{Truncation selection} where we rank all agents in the population by episodic reward. If the current agent is in the bottom 20\% of the population, we sample another agent uniformly from the top 20\% of the population, and copy its weights and hyperparameters. 
\vspace{-0.5em}\paragraph{Explore}We consider two exploration strategies in hyperparameter space. (a) \emph{Perturb}, where each hyperparameter independently is randomly perturbed by a factor of 1.2 or 0.8. (b) \emph{Resample}, where each hyperparameter is resampled from the original prior distribution defined with some probability.

\subsubsection{Results}

\begin{figure}[htb]
\centering
\begin{tabular}{ll}
\includegraphics[width=0.45\textwidth]{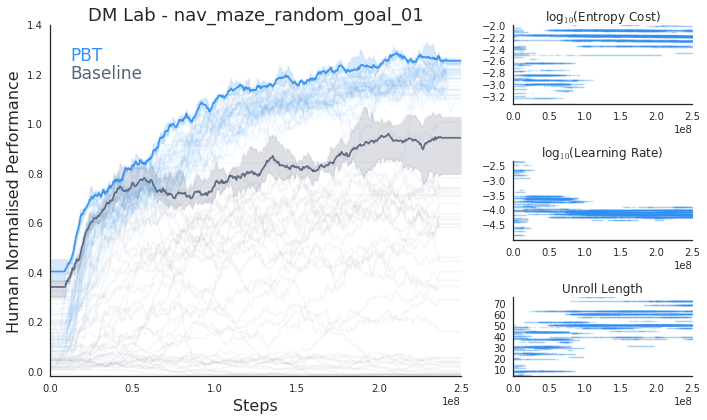}&
\includegraphics[width=0.45\textwidth]{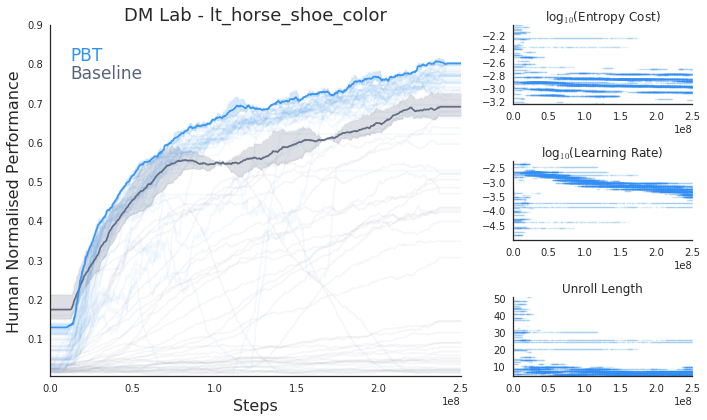}\\
\includegraphics[width=0.45\textwidth]{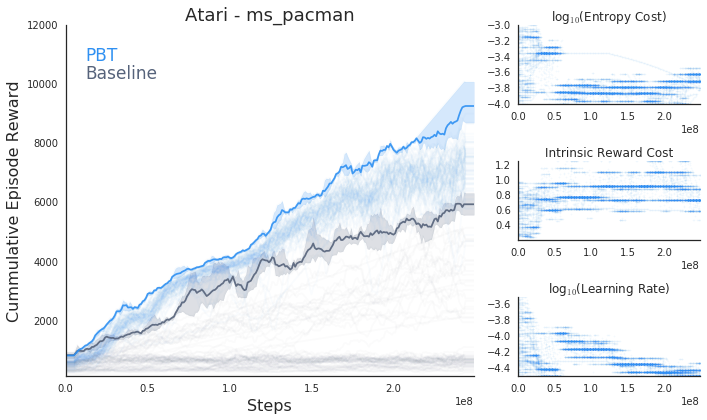}&
\includegraphics[width=0.45\textwidth]{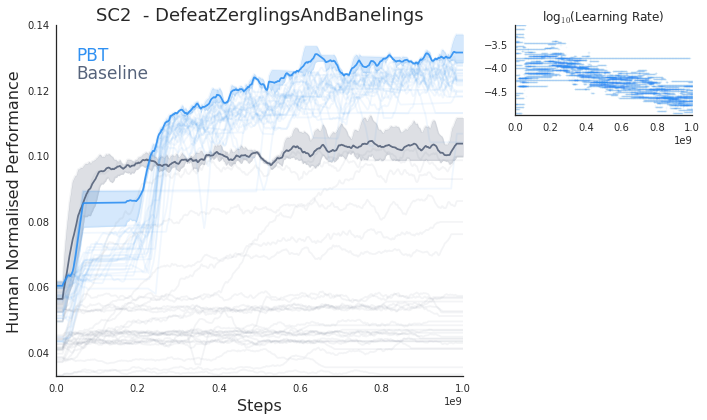}\\
\includegraphics[width=0.45\textwidth]{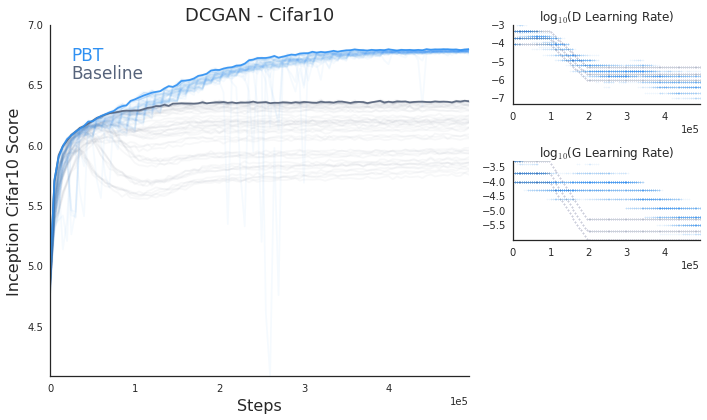}&
\includegraphics[width=0.45\textwidth]{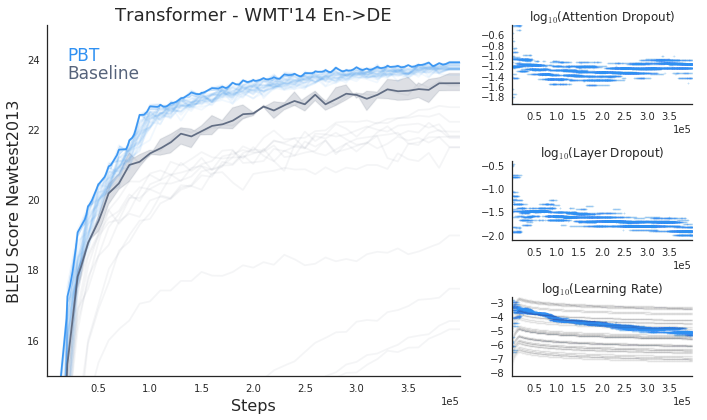}
\end{tabular}
\caption{\small Training curves of populations trained with PBT (blue) and without PBT (black) for individual levels on DeepMind Lab (DM Lab), Atari, and Starcraft~II (SC2), as well as GAN training on CIFAR-10 (CIFAR-10 Inception score is plotted as this is what is optimised for by PBT). Faint lines represent each individual worker, while thick lines are an average of the top five members of the population at each timestep. The hyperparameter populations are displayed to the right, and show how the populations' hyperparameters adapt throughout training, including the automatic discovery of learning rate decay and, for example, the discovery of the benefits of large unroll lengths on DM Lab.
}
\label{fig:curves}
\end{figure}

\begin{figure}[t]
\centering
\begin{tabular}{cccc}
\includegraphics[width=0.8\textwidth]{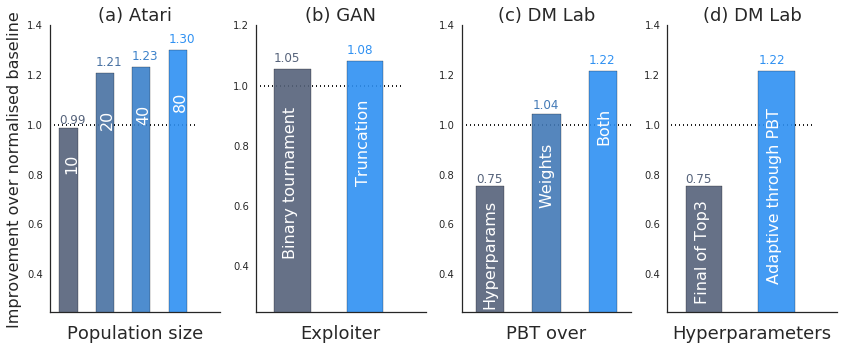}
\end{tabular}
\caption{\small A highlight of results of various ablation studies on the design space of PBT, reported as improvement over the equivalent random search baseline. (a)~\emph{Population size:} We evaluate the effect of population size on the performance of PBT when training FuN on Atari. In general we find that a smaller population sizes leads to higher variance and suboptimal results, however using a population of 20 and over achieves consistent improvements, with diminishing returns for larger populations. (b)~\emph{Exploiter:} We compare the effect of using different \pbtexploit functions -- binary tournament and truncation selection. When training GANs, we find truncation selection to work better. (c)~\emph{PBT over:} We show the effect of removing parts of PBT when training UNREAL on DeepMind Lab, in particular only performing PBT on hyperparameters (so model weights are not copied between workers during \pbtexploit), and only performing PBT on model weights (so hyperparameters are not copied between workers or explored). We find that indeed it is the combination of optimising hyperparameters as well as model selection which leads to best performance. (d)~\emph{Hyperparameters:} Since PBT allows online adaptation of hyperparameters during training, we evaluate how important the adaptation is by comparing full PBT performance compared to using the set of hyperparameters that PBT found by the end of training. When training UNREAL on DeepMind Lab we find that the strength of PBT is in allowing the hyperparameters to be adaptive, not merely in finding a good prior on the space of hyperparameters.
}
\label{fig:ablation}
\end{figure}

The high level results are summarised in \figref{fig:showstopper}. On all three domains -- DeepMind Lab, Atari, and StarCraft~II -- PBT increases the final performance of the agents when trained for the same number of steps, compared to the very strong baseline of performing random search with the same number of workers. 

On DeepMind Lab, when UNREAL is trained using a population of 40 workers, using PBT increases the final performance of UNREAL from 93\% to 106\% human performance, when performance is averaged across all levels. This represents a significant improvement over an already extremely strong baseline, with some levels such as $\mathrm{nav\_maze\_random\_goal\_01}$ and $\mathrm{lt\_horse\_shoe\_color}$ showing large gains in performance due to PBT, as shown in \figref{fig:curves}. PBT seems to help in two main ways: first is that the hyperparameters are clearly being focused on the best part of the sampling range, and adapted over time. For example, on $\mathrm{lt\_horse\_shoe\_color}$ in \figref{fig:curves} one can see the learning rate being annealed by PBT as training progresses, and on $\mathrm{nav\_maze\_random\_goal\_01}$ in \figref{fig:curves} one can see the unroll length -- the number of steps the agent (a recurrent neural network) is unrolled before performing N-step policy gradient and backpropagation through time -- is gradually increased. This allows the agent to learn to utilise the recurrent neural network for memory, allowing it to obtain superhuman performance on $\mathrm{nav\_maze\_random\_goal\_01}$ which is a level which requires memorising the location of a goal in a maze, more details can be found in~\citep{jaderberg2016reinforcement}. Secondly, since PBT is copying the weights of good performing agents during the exploitation phase, agents which are lucky in environment exploration are quickly propagated to more workers, meaning that all members of the population benefit from the exploration luck of the remainder of the population.

When we apply PBT to training FuN on 4 Atari levels, we similarly see a boost in final performance, with PBT increasing the average human normalised performance from 147\% to 181\%. On particular levels, such as $\mathrm{ms\_pacman}$ PBT increases performance significantly from 6506 to 9001 (\figref{fig:curves}), which represents state-of-the-art for an agent receiving only pixels as input. 

Finally on StarCraft II, we similarly see PBT improving A3C baselines from 36\% huamn performance to 39\% human performance when averaged over 6 levels, benefiting from automatic online learning rate adaptation and model selection such as shown in \figref{fig:curves}.

More detailed experimental results, and details about the specific experimental protocols can be found in Appendix \ref{sec:detailed_results_RL}.

\subsection{Machine Translation}\label{sec:language}

\begin{figure}[t]
\centering
\includegraphics[width=0.475\textwidth]{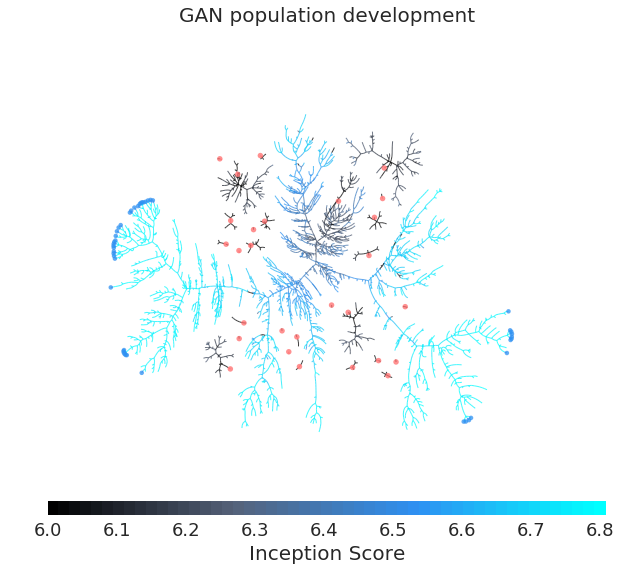}
\includegraphics[width=0.475\textwidth]{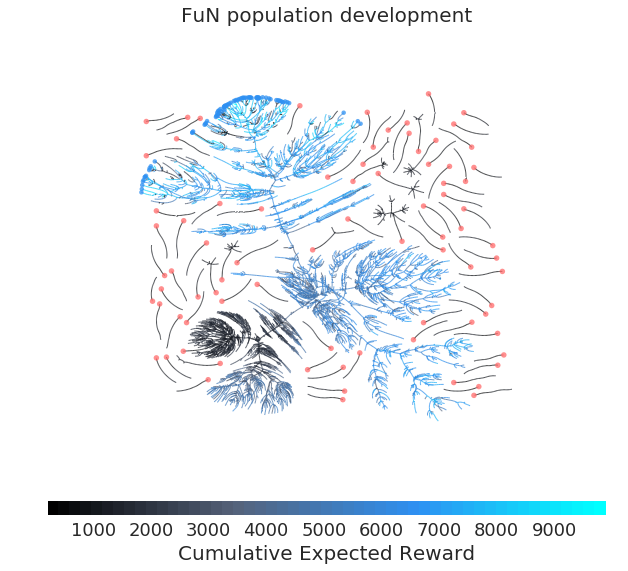}
\caption{\small Model development analysis for GAN (left) and Atari (right) experiments -- full phylogenetic tree of entire training. Pink dots represent initial agents, blue ones the final ones. Branching of the graph means that the exploit operation has been executed (and so parameters were copied), while paths represent consecutive updates using the step function. Colour of the edges encode performance, from black (weak model) to cyan (strong model).
}
\label{fig:trees}
\end{figure}

As an example of applying PBT on a supervised learning problem, we look at training neural machine translation models. In this scenario, the task is to encode a source sequence of words in one language, and output a sequence in a different target language. We focus on English to German translation on the WMT 2014 English-to-German dataset, and use a state-of-the-art model, Transformer networks~\citep{vaswani2017attention}, as the baseline model to apply PBT to. Crucially, we also use the highly optimised learning rate schedules and hyperparameter values for our baselines to compare to -- these are the result of a huge amount of hand tuning and Bayesian optimisation.

\subsubsection{PBT for Machine Translation}
We use a population size of 32 workers, where each member of the population is a single GPU optimising a Transformer network for $400\times10^3$ steps. 
\vspace{-0.5em}\paragraph{Hyperparameters}We allow PBT to optimise the learning rate, attention dropout, layer dropout, and ReLU dropout rates. 
\vspace{-0.5em}\paragraph{Step}Each step does a step of gradient descent with Adam~\citep{adam}. 
\vspace{-0.5em}\paragraph{Eval}We evaluate the current model by computing the BLEU score on the WMT \textit{newstest2012} dataset. This allows us to optimise our population for BLEU score, which usually cannot be optimised for directly. 
\vspace{-0.5em}\paragraph{Ready}A member of the population is deemed ready to exploit and explore using the rest of the population every $2\times10^3$ steps. 
\vspace{-0.5em}\paragraph{Exploit}We use the \emph{t-test selection} exploitation method described in \sref{sec:pbtforrl}. \vspace{-0.5em}\paragraph{Explore}We explore hyperparameter space by the \emph{perturb} strategy described in~\sref{sec:pbtforrl} with 1.2 and 0.8 perturbation factors.

\subsubsection{Results}

\begin{figure}[t]
\centering
\includegraphics[width=0.8\textwidth]{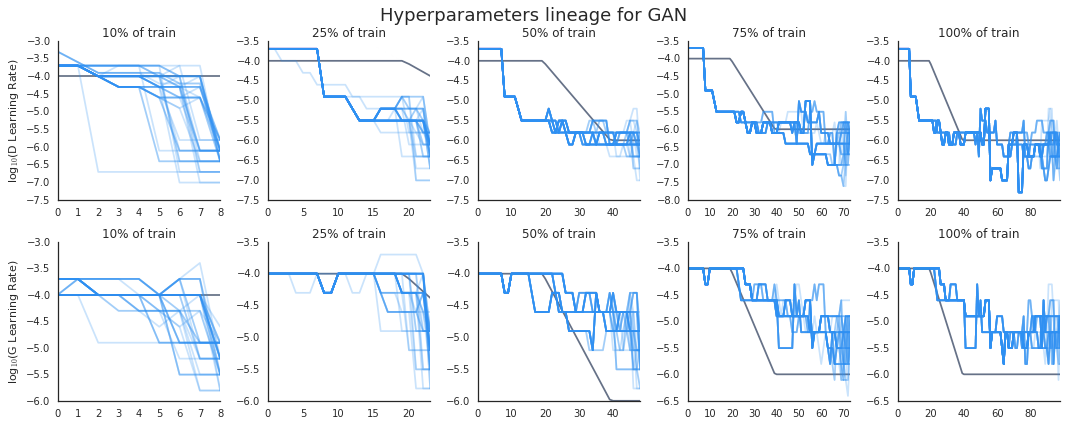}\\
\includegraphics[width=0.8\textwidth]{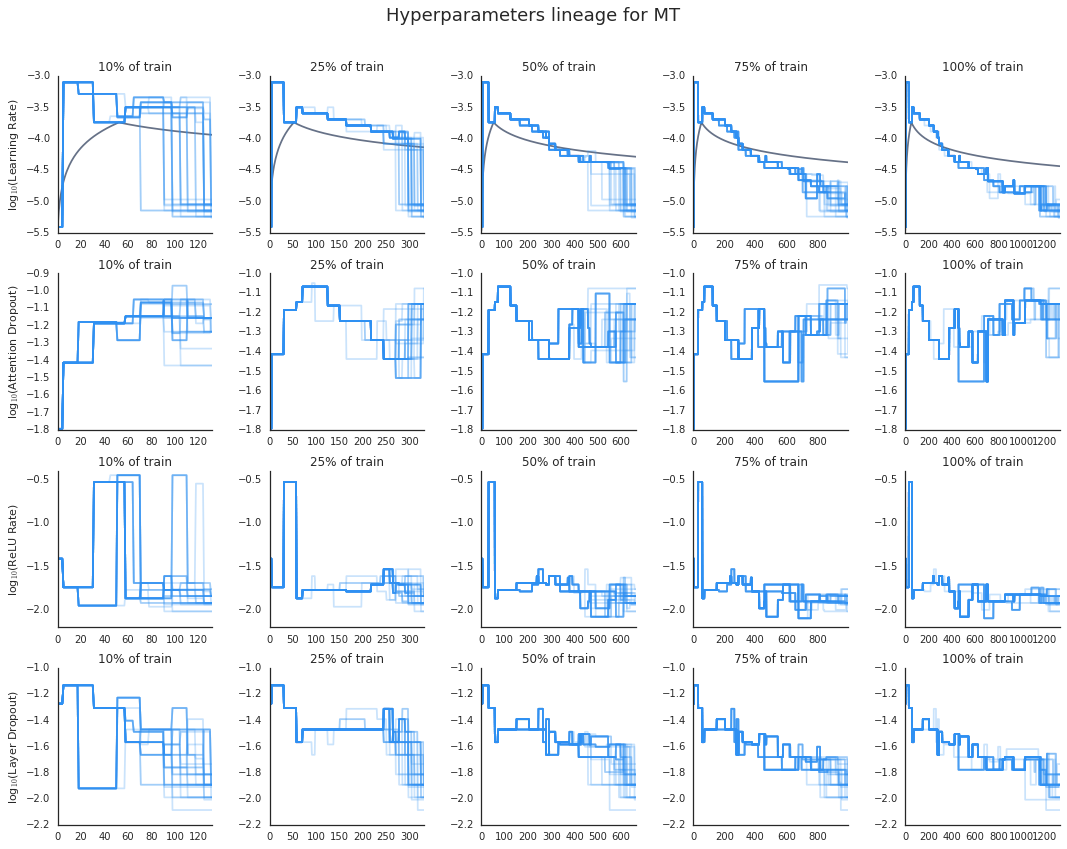}
\caption{\small Agent development analysis for GAN (top) and machine translation (bottom) experiments -- lineages of hyperparameters changes for the final agents after given amount of training. Black lines show the annealing schedule used by the best baseline run.
}
\label{fig:lineages}
\end{figure}

When we apply PBT to the training of Transformer networks, we find that PBT actually results in models which exceed the performance of this highly tuned baseline model: BLEU score on the validation set (\textit{newstest2012}) is improved from 23.71 to 24.23, and on the test set of WMT 2014 English-to-German is improved from 22.30 to 22.65 using PBT (here we report results using the small Transformer network using a reduced batch size, so is not representative of state of the art). PBT is able to automatically discover adaptations of various dropout rates throughout training, and a learning rate schedule that remarkably resembles that of the hand tuned baseline: very initially the learning rate starts small before jumping by three orders of magnitude, followed by something resembling an exponential decay, as shown in the lineage plot of \figref{fig:lineages}. We can also see from \figref{fig:curves} that the PBT model trains much faster.

\subsection{Generative Adversarial Networks}\label{sec:gans}
The Generative Adversarial Networks (GAN)~\citep{goodfellowgan} framework
learns generative models via a training paradigm consisting of two competing modules --
a \emph{generator} and a \emph{discriminator} (alternatively called a \emph{critic}).
The discriminator takes as input samples from the generator and the real data distribution,
and is trained to predict whether inputs are real or generated.
The generator typically maps samples from a simple noise distribution to a complex data distribution (\eg~images) with the objective of maximally fooling the discriminator into classifying or scoring its samples as real.

Like RL agent training, GAN training can be remarkably brittle and unstable in the face of suboptimal hyperparameter selection and even unlucky random initialisation,
with generators often collapsing to a single mode or diverging entirely.
Exacerbating these difficulties, the presence of two modules optimising competing objectives doubles the number of hyperparameters,
often forcing researchers to choose the same hyperparameter values for both modules.
Given the complex learning dynamics of GANs,
this unnecessary coupling of hyperparameters between the two modules is unlikely to be the best choice.

Finally, the various metrics used by the community to evaluate the quality of samples
produced by GAN generators are necessarily distinct from those used for the adversarial optimisation itself.
We explore whether we can improve the performance of generators under these metrics by directly targeting them
as the PBT meta-optimisation evaluation criteria.

\subsubsection{PBT for GANs}
We train a population of size 45 with each worker being trained for $10^6$ steps. 
\vspace{-0.5em}\paragraph{Hyperparameters}We allow PBT to optimise the discriminator's learning rate and the generator's learning rate separately. 
\vspace{-0.5em}\paragraph{Step}Each step consists of $K=5$ gradient descent updates of the discriminator followed by a single update of the generator
using the Adam optimiser~\citep{adam}.
We train using the WGAN-GP~\citep{improvedwgan} objective where the discriminator estimates the Wasserstein distance between the real and generated data distributions, with the Lipschitz constraint enforced by regularising its input gradient to have a unit norm.
See Appendix~\ref{sec:gan_details} for additional details. 
\vspace{-0.5em}\paragraph{Eval}We evaluate the current model by a variant of the Inception score proposed by~\citet{openaigan}
computed from the outputs of a pretrained CIFAR classifier (as used in~\citet{mihaelaalphagan})
rather than an ImageNet classifier,
to avoid directly optimising the final performance metric.
The CIFAR Inception scorer uses a much smaller network, making evaluation much faster. 
\vspace{-0.5em}\paragraph{Ready}A member of the population is deemed ready to exploit and explore using the rest of the population every $5\times10^3$ steps. 
\vspace{-0.5em}\paragraph{Exploit}We consider two exploitation strategies.
(a) \emph{Truncation selection} as described in~\sref{sec:pbtforrl}.
(b) \emph{Binary tournament} in which each member of the population randomly selects another member of the population, and copies its parameters if the other member's score is better. Whenever one member of the population is copied to another, all parameters -- the hyperparameters, weights of the generator, and weights of the discriminator -- are copied. 
\vspace{-0.5em}\paragraph{Explore}We explore hyperparameter space by the \emph{perturb} strategy described in~\sref{sec:pbtforrl}, but with more aggressive perturbation factors of 2.0 or 0.5.

\subsubsection{Results}
We apply PBT to optimise the learning rates for a CIFAR-trained GAN discriminator and generator
using architectures similar to those used DCGAN~\citep{dcgan}.
For both PBT and the baseline, the population size is $45$, with learning rates initialised to
$\{1, 2, 5\}\times10^{-4}$
for a total of $3^2 = 9$ initial hyperparameter settings,
each replicated $5$ times with independent random weight initialisations.
Additional architectural and hyperparameter details can be found in Appendix~\ref{sec:gan_details}.
The baseline utilises an exponential learning rate decay schedule, which we found to work the best among a number of hand-designed annealing strategies, detailed in Appendix~\ref{sec:gan_details}.

In~\figref{fig:ablation} (b), we compare the \emph{truncation selection} exploration strategy with a simpler variant, \emph{binary tournament}.
We find a slight advantage for truncation selection,
and our remaining discussion focuses on results obtained by this strategy.
\figref{fig:curves} (bottom left) plots the CIFAR Inception scores and shows how the learning rates of the population change throughout the first half of training (both the learning curves and hyperparameters saturate and remain roughly constant in the latter half of training).
We observe that PBT automatically learns to decay learning rates,
though it does so dynamically, with irregular flattening and steepening, resulting in a
learnt schedule that would not be matched by any typical exponential or linear decay protocol.
Interestingly, the learned annealing schedule for the discriminator's learning rate closely tracks that of the best performing baseline (shown in grey), while the generator's schedule deviates from the baseline schedule substantially.

In Table~\ref{tab:ganresults} (Appendix~\ref{sec:gan_details}),
we report standard ImageNet Inception scores as our main performance measure, but use CIFAR Inception score for model selection, as optimising ImageNet Inception score directly would potentially overfit our models to the test metric.
CIFAR Inception score is used to select the best model after training is finished both for the baseline and for PBT.
Our best-performing baseline achieves a CIFAR Inception score of $6.39$, corresponding to an ImageNet Inception score of $6.45
$, similar to or better than Inception scores for DCGAN-like architectures reported in prior work~\citep{mihaelaalphagan,improvedwgan,lrgan}.
Using PBT, we strongly outperform this baseline, achieving a CIFAR Inception score of $6.80$ and corresponding ImageNet Inception score of $6.89
$.
For reference,
Table~\ref{tab:ganresults} (Appendix~\ref{sec:gan_details}) also reports the peak ImageNet Inception scores obtained by any population member at any point during training.
\figref{fig:gan_samples} shows samples from the best generators with and without PBT.

\subsection{Analysis}\label{sec:analysis}

In this section we analyse and discuss the effects of some of the design choices in setting up PBT, as well as probing some aspects of why the method is so successful.

\paragraph{Population Size}\label{sec:analysis_pop_size}

In \figref{fig:ablation}~(a) we demonstrate the effect of population size on the performance of PBT when training FuN on Atari. In general, we find that if the population size is too small (10 or below) we tend to encounter higher variance and can suffer from poorer results -- this is to be expected as PBT is a greedy algorithm and so can get stuck in local optima if there is not sufficient population to maintain diversity and scope for exploration. However, these problems rapidly disappear as population size increases and we see improved results as the population size grows. In our experiments, we observe that a population size of between $20$ and $40$ is sufficient to see strong and consistent improvements; larger populations tend to fare even better, although we see diminishing returns for the cost of additional population members.

\paragraph{Exploitation Type}\label{sec:analysis_exploit}

In \figref{fig:ablation}~(b) we compare the effect of using different \pbtexploit functions -- binary tournament and truncation selection. For GAN training, we find truncation selection to work better. Similar behaviour was seen in our preliminary experiments with FuN on Atari as well.
Further work is required to fully understand the impact of different \pbtexploit mechanisms, however it appears that truncation selection is a robustly useful choice.

\paragraph{PBT Targets}\label{sec:analysis_pbt_targets}

In \figref{fig:ablation}~(c) we show the effect of only applying PBT to subsets of parameters when training UNREAL on DM Lab. In particular, we contrast (i) only performing exploit-and-explore on hyperparameters (so model weights are not copied between workers during \pbtexploit); and (ii) only performing exploit-and-explore on model weights (so hyperparameters are not copied between workers or explored). Whilst there is some increase in performance simply from the selection and propagation of good model weights, our results clearly demonstrate that in this setting it is indeed the \emph{combination} of optimising hyperparameters as well as model selection that lead to our best performance.

\paragraph{Hyperparameter Adaptivity}\label{sec:analysis_hyper_adapt}

In \figref{fig:ablation}~(d) we provide another demonstration that the benefits of PBT go beyond simply finding a single good fixed hyperparameter combination. Since PBT allows online adaptation of hyperparameters during training, we evaluate how important the adaptation is by comparing full PBT performance compared to using the set of hyperparameters that PBT found by the end of training. When training UNREAL on DeepMind Lab we find that the strength of PBT is in allowing the hyperparameters to be adaptive, not merely in finding a good prior on the space of hyperparameters

\paragraph{Lineages}\label{sec:analysis_lineage}

\figref{fig:lineages} shows hyperparameter sequences for the final agents in the population for GAN and Machine Translation (MT) experiments. We have chosen these two examples as the community has developed very specific learning rate schedules for these models. In the GAN case, the typical schedule involves linearly annealed learning rates of both generator and discriminator in a synchronised way. As one can see, PBT discovered a similar scheme, however the discriminator annealing is much more severe, while the generator schedule is slightly less aggressive than the typical one. 

As mentioned previously, the highly non-standard learning rate schedule used for machine translation has been to some extent replicated by PBT -- but again it seems to be more aggressive than the hand crafted version. Quite interestingly, the dropout regime is also non-monotonic, and follows an even more complex path -- it first increases by almost an order of magnitude, then drops back to the initial value, to slowly crawl back to an order of magnitude bigger value. These kind of schedules seem easy to discover by PBT, while at the same time are beyond the space of hyperparameter schedules considered by human experts.

\paragraph{Phylogenetic trees}\label{sec:analysis_phylogenetic}

Figure \ref{fig:trees} shows a phylogenetic forest for the population in the GAN and Atari experiments. A property made noticeable by such plots is that all final agents are descendants of the same, single initial agent. This is on one hand a consequence of the greedy nature of specific instance of PBT used, as \pbtexploit only uses a single other agent from the population. In such scenario, the number of original agents can only decrease over time, and so should finally converge to the situation observed. On the other hand one can notice that in both cases there are quite rich sub-populations which survived for a long time, showing that the PBT process considered multiple promising solutions before finally converging on a single winner -- an exploratory capability most typical optimisers lack.

One can also note the following interesting qualitative differences between the two plots: in the GAN example, almost all members of the population enjoy an improvement in score relative to their parent; however, in the Atari experiment we note that even quite late in learning there can be exploration steps that result in a sharp drop in cumulative episode reward relative to the parent. This is perhaps indicative of some instability of neural network based RL on complex problems -- an instability that PBT is well positioned to correct.

\section{Conclusions}\label{sec:conclusion}

We have presented Population Based Training, which represents a practical way to augment the standard training of neural network models. We have shown consistent improvements in accuracy, training time and stability across a wide range of domains by being able to optimise over weights and hyperparameters jointly. It is important to note that contrary to conventional hyperparameter optimisation techniques, PBT discovers an adaptive schedule rather than a fixed set of hyperparameters. We already observed significant improvements on a wide range of challenging problems including state of the art models on deep reinforcement learning and hierarchical reinforcement learning, machine translation, and GANs. While there are many improvements and extensions to be explored going forward, we believe that the ability of PBT to enhance the optimisation process of new, unfamiliar models, to adapt to non-stationary learning problems, and to incorporate the optimisation of indirect performance metrics and auxiliary tasks, results in a powerful platform to propel future research.

\subsubsection*{Acknowledgments}
We would like to thank Yaroslav Ganin, Mihaela Rosca, John Agapiou, Sasha Vehznevets, Vlad Firoiu, and the wider DeepMind team for many insightful discussions, ideas, and support.

\bibliography{iclr2018_conference}
\bibliographystyle{iclr2018_conference}

\begin{appendices}

\newpage
\section{Population Based Training of Neural Networks Appendix}\label{sec:appendix}

\subsection{Practical implementations}\label{sec:practicum}

In this section, we describe at a high level how one might concretely implement PBT on a commodity training platform.
Relative to a standard setup in which multiple hyperparameter configurations are searched in parallel
(\eg~grid/random search), a practitioner need only add the ability for population members to read and write to a shared data-store (\eg~a key-value store, or a simple file-system). Augmented with such a setup, there are two main types of interaction required:

(1)  Members of the population interact with this data-store to update their current performance and can also use it to query the recent performance of other population members.

(2)  Population members periodically checkpoint themselves, and when they do so they write their performance to the shared data-store. In the exploit-and-explore process, another member of the population may elect to restore its weights from the checkpoint of another population member (\ie~exploiting), and would then modify the restored hyperparameters (\ie~exploring) before continuing to train.

As a final comment, although we present and recommend PBT as an asynchronous parallel algorithm, it could also be executed  semi-serially or with partial synchrony. 
Although it would obviously result in longer training times than running fully asynchronously in parallel, one would still gain the performance-per-total-compute benefits of PBT relative to random-search. This execution mode may be attractive to researchers who wish to take advantage of commodity compute platforms at cheaper, but less reliable, preemptible/spot tiers.
As an example of an approach with partial synchrony, one could imagine the following procedure: each member of the population is advanced in parallel (but without synchrony) for a fixed number of steps;  once all (or some large fraction) of the population has advanced, one would call the exploit-and-explore procedure; and then repeat. This partial locking would be forgiving of different population members running at widely varying speeds due to preemption events, etc.

\subsection{Detailed Results: RL}\label{sec:detailed_results_RL}

\subsubsection*{FuN - Atari}

Table \ref{tab:FuN_DetailedResults} shows the breakdown of results per game for FuN on the Atari domain, with and without PBT for different population sizes. The initial hyperparameter ranges for the population in both the FuN and PBT-FuN settings were the same as those used for the corresponding games in the original Feudal Networks paper ~\citep{vezhnevets2017feudal}, with the exception of intrinsic-reward cost (which was sampled in the range $[0.25, 1]$ , rather than $[0,1]$ -- since we have found that very low intrinsic reward cost weights are rarely effective). We used fixed discounts of $0.95$ in the worker and $0.99$ in the manager since, for the games selected, these were shown in the original FuN paper to give good performance. The subset of games were themselves chosen as ones for which the hierarchical abstractions provided by FuN give particularly strong performance improvements over a flat A3C agent baseline. In these experiments, we used \emph{Truncation} as the exploitation mechanism, and \emph{Perturb} as the exploration mechanism. Training was performed for $2.5\times 10^8$ agent-environmnent interactions, corresponding to $1\times 10^9$ total environment steps (due to use of action-repeat of $4$). Each experiment condition was repeated $3$ times, and the numbers in the table reflect the mean-across-replicas of the single best performing population member at the end of the training run.

\begin{table}[h]
\scriptsize
    \begin{center}
    \begin{tabular}
    {lcccccccc}
    \toprule
      \textbf{Game}  & \multicolumn{2}{c}{\textbf{Pop $=$ 10 }} &\multicolumn{2}{c}{\textbf{Pop $=$ 20 }} & \multicolumn{2}{c}{\textbf{Pop $=$ 40 }} & \multicolumn{2}{c}{\textbf{Pop $=$ 80 }} \\ 
    \midrule
    & \textbf{FuN} & \textbf{PBT-FuN}  & \textbf{FuN} & \textbf{PBT-FuN}  & \textbf{FuN} & \textbf{PBT-FuN}  & \textbf{FuN} & \textbf{PBT-FuN} \\ 
    
    amidar & 1742 {\tiny $\pm$174} & 1511 {\tiny $\pm$400} & 1645 {\tiny $\pm$158} & 2454 {\tiny $\pm$160} & 1947 {\tiny $\pm$80} & 2752 {\tiny $\pm$81} & 2055 {\tiny $\pm$113} & 3018 {\tiny $\pm$269} \\ 
    gravitar & 3970 {\tiny $\pm$145} & 4110 {\tiny $\pm$225} & 3884 {\tiny $\pm$362} & 4202 {\tiny $\pm$113} & 4018 {\tiny $\pm$180} & 4481 {\tiny $\pm$540} & 4279 {\tiny $\pm$161} & 5386 {\tiny $\pm$306} \\
    ms-pacman & 5635 {\tiny $\pm$656} & 5509 {\tiny $\pm$589} & 5280 {\tiny $\pm$349} & 5714 {\tiny $\pm$347} & 6177 {\tiny $\pm$402} & 8124 {\tiny $\pm$392} & 6506 {\tiny $\pm$354} & 9001 {\tiny $\pm$711} \\
    enduro & 1838 {\tiny $\pm$37} & 1958 {\tiny $\pm$81} & 1881 {\tiny $\pm$63} & 2213 {\tiny $\pm$6} & 2079 {\tiny $\pm$9} & 2228 {\tiny $\pm$77} & 2166 {\tiny $\pm$34} & 2292 {\tiny $\pm$16} \\
    \bottomrule

    \end{tabular}
    \end{center}
    \caption{\footnotesize Per game results for Feudal Networks (FuN) with and without PBT for different population sizes (Pop). }
    \label{tab:FuN_DetailedResults}
\end{table}

\subsubsection*{UNREAL - DM Lab}

Table \ref{tab:UNREAL_DetailedResults} shows the breakdown on results per level for UNREAL and PBT-UNREAL on the DM Lab domain. The following three hyperparameters were randomly sampled for both methods and subsequently perturbed in the case of PBT-UNREAL: the learning rate was sampled in the log-uniform range [0.00001, 0.005], the entropy cost was sampled in the log-uniform range [0.0005, 0.01], and the unroll length was sampled uniformly between 5 and 50 steps. Other experimental details were replicated from \cite{jaderberg2016reinforcement}, except for pixel control weight which was set to 0.1.

\begin{table}[h]
\scriptsize
    \begin{center}
    \begin{tabular}
    {lcc|cccc}
    \toprule
      \textbf{Game}  & \multicolumn{2}{c}{\textbf{Pop $=$ 40 }} &  \multicolumn{4}{c}{\textbf{Pop $=$ 20}}\\ 
    \midrule
    & \textbf{UNREAL} & \textbf{PBT-UNREAL} &  \textbf{UNREAL} & \textbf{PBT-Hypers-UNREAL} & \textbf{PBT-Weights-UNREAL} & \textbf{PBT-UNREAL}\\
    
    emstm\_non\_match & 66.0 & 66.0 & 42.6 & 14.5 & 56.0 & 53.2 \\
    emstm\_watermaze & 39.0 & 36.7 & 27.9 & 33.2 & 34.6 & 36.8 \\
    lt\_horse\_shoe\_color & 75.4 & 90.2 & 77.4 & 44.2 & 56.3 & 87.5\\
    nav\_maze\_random\_goal\_01 & 114.2 & 149.4 & 82.2 & 81.5 & 74.2 & 96.1\\
    lt\_hallway\_slope & 90.5 & 109.7 \\
    nav\_maze\_all\_random\_01 & 59.6 & 78.1 \\
    nav\_maze\_all\_random\_02 & 62.0 & 92.4 \\
    nav\_maze\_all\_random\_03 & 92.7 & 115.3 \\
    nav\_maze\_random\_goal\_02 & 145.4 & 173.3 \\
    nav\_maze\_random\_goal\_03 & 127.3 & 153.5 \\
    nav\_maze\_static\_01 & 131.5 & 133.5 \\
    nav\_maze\_static\_02 & 194.5 & 220.2 \\
    nav\_maze\_static\_03 & 588.9 & 594.8 \\
    seekavoid\_arena\_01 & 42.5 & 42.3 \\
    stairway\_to\_melon\_01 & 189.7 & 192.4 \\
    \bottomrule

    \end{tabular}
    \end{center}
    \caption{\footnotesize Per game results for UNREAL with and without PBT.}
    \label{tab:UNREAL_DetailedResults}
\end{table}

\subsubsection*{A3C - StarCraft II}

Table \ref{tab:UNREAL_DetailedResults} shows the breakdown on results per level for UNREAL on the DM Lab domain, with and without PBT for different population sizes. We take the baseline setup of \cite{vinyals2017starcraft} with feed-forward network agents, and sample the learning rate from the log-uniform range [0, 0.001].

\begin{table}[h]
\scriptsize
    \begin{center}
    \begin{tabular}
    {lcc}
    \toprule
      \textbf{Game}  & \multicolumn{2}{c}{\textbf{Pop $=$ 30 }} \\ 
    \midrule
    & \textbf{A3C} & \textbf{PBT-A3C} \\
    
    CollectMineralShards & 93.7 & 100.5 \\
    FindAndDefeatZerglings & 46.0 & 49.9 \\
    DefeatRoaches & 121.8 & 131.5 \\
    DefeatZerglingsAndBanelings & 94.6 & 124.7 \\
    CollectMineralsAndGas & 3345.8 & 3345.3 \\
    BuildMarines & 0.17 & 0.37 \\
    \bottomrule

    \end{tabular}
    \end{center}
    \caption{\footnotesize Per game results for A3C on StarCraft II with and without PBT.}
    \label{tab:SC_DetailedResults}
\end{table}

\subsection{Detailed Results: GAN}
\label{sec:gan_details}

\begin{figure}
\centering

\includegraphics[width=0.75\textwidth,trim={0 2.296cm 0 0},clip]{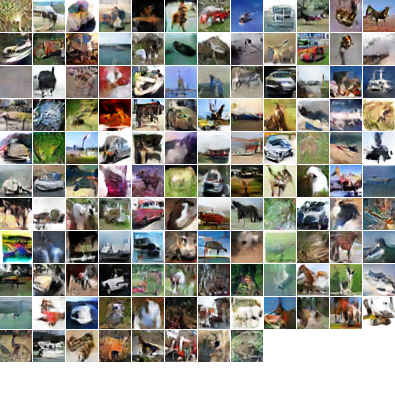} \\
Baseline GAN
\\
\vspace{1cm}
\includegraphics[width=0.75\textwidth,trim={0 2.296cm 0 0},clip]{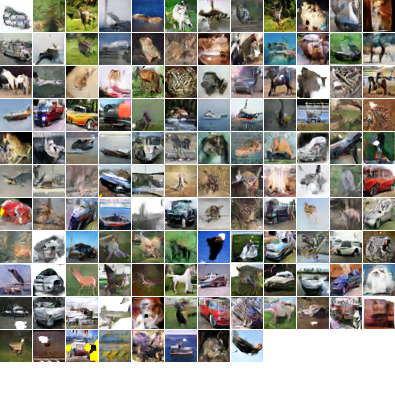} \\
PBT-GAN
\caption{CIFAR GAN samples generated by the best performing generator of the Baseline GAN (top) and PBT-GAN (bottom).}
\label{fig:gan_samples}
\end{figure}

This section provides additional details of and results from our GAN experiments introduced in~\sref{sec:gans}.
\figref{fig:gan_samples} shows samples generated by the best performing generators with and without PBT.
In Table~\ref{tab:ganresults}, we report GAN Inception scores with and without PBT.

\paragraph{Architecture and hyperparameters}

The architecture of the discriminator and the generator are similar to those proposed in DCGAN~\citep{dcgan}, but with $4\times4$ convolutions (rather than $5\times5$), and a smaller generator with half the number of feature maps in each layer.
Batch normalisation is used in the generator, but not in the discriminator, as in~\citet{improvedwgan}.
Optimisation is performed using the Adam optimiser~\citep{adam}, with $\beta_1=0.5$, $\beta_2=0.999$, and $\epsilon=10^{-8}$.
The batch size is 128.

\begin{table}
    \begin{center}
    \begin{tabular}{lccc}
    \toprule
                                  & \bf{GAN}           & \bf{PBT-GAN (TS)}  & \bf{PBT-GAN (BT)}  \\
    \midrule
    CIFAR Inception               & 6.38$\pm$0.03 & 6.80$\pm$0.01 & 6.77$\pm$0.02          \\
    ImageNet Inception (by CIFAR) & 6.45$\pm$0.05 & 6.89$\pm$0.04 & 6.74$\pm$0.05 \\
    ImageNet Inception (peak)     & 6.48$\pm$0.05 & 7.02$\pm$0.07 & 6.81$\pm$0.04 \\
    \bottomrule
    \end{tabular}
    \end{center}
    \caption{CIFAR and ImageNet Inception scores for GANs trained with and without PBT, using the \emph{truncation selection} (TS) and \emph{binary tournament} (BT) exploitation strategies.
    The \emph{ImageNet Inception (by CIFAR)} row reports the ImageNet Inception score for the best generator (according to the CIFAR Inception score) after training is finished, while \emph{ImageNet Inception (peak)} gives the best scores (according to the ImageNet Inception score) obtained by any generator at any point during training.}
    \label{tab:ganresults}
\end{table}

\paragraph{Baseline selection}
To establish a strong and directly comparable baseline for GAN PBT, we extensively searched over and chose the best by CIFAR Inception score among many learning rate annealing strategies used in the literature.
For each strategy, we trained for a total of $2N$ steps, where learning rates were fixed to their initial settings for the first $N$ steps, then for the latter $N$ steps either (a) exponentially decayed to $10^{-2}$ times the initial setting, or (b) linearly decayed to $0$,
searching over $N=\{0.5, 1, 2\}\times10^5$.
(Searching over the number of steps $N$ is essential due to the instability of GAN training: performance often degrades with additional training.)
In each training run to evaluate a given combination of annealing strategy and choice of training steps $N$,
we used the same number of independent workers ($45$) and initialised the learning rates to the same values used in PBT.
The best annealing strategy we found using this procedure was exponential decay with $N=1\times10^5$ steps.
For the final baseline, we then extended this strategy to train for the same number of steps as used in PBT ($10^6$) by continuing training for the remaining $8\times10^5$ steps with constant learning rates fixed to their final values after exponential decay.

\subsection{Detailed Results: Machine Translation}
\label{sec:mt_baseline}
We used the open-sourced implementation of the Transformer framework\footnote{https://github.com/tensorflow/tensor2tensor} with the provided \texttt{transformer\_base\_single\_gpu} architecture settings. This model has the same number of parameters as the base configuration that runs on 8 GPUs ($6.5\times10^7$), but sees, in each training step, $\frac{1}{16}$ of the number of tokens ($2048$ vs. $8 \times 4096$) as it uses a smaller batch size. For both evolution and the random-search baseline, we searched over learning rate, and dropout values applied to the attention softmax activations, the outputs of the ReLU units in the feed-forward modules, and the output of each layer. We left other hyperparameters unchanged and made no other changes to the implementation. For evaluation, we computed BLEU score based on auto-regressive decoding with beam-search of width 4. We used \textit{newstest2012} and \textit{newstest2013} respectively as the evaluation set used by PBT, and as the test set for monitoring. At the end of training, we report tokenized BLEU score on \textit{newstest2014} as computed by \texttt{multi-bleu.pl} script\footnote{https://github.com/moses-smt/mosesdecoder/blob/master/scripts/generic/multi-bleu.perl}. We also evaluated the original hyperparameter configuration trained for the same number of steps and obtained the BLEU score of $21.23$, which is lower than both our baselines and PBT results.

\end{appendices}

\end{document}